%% file: Multilayertransformlearning.tex
\documentclass{article}
\usepackage{spconf}

\usepackage{cite}
\usepackage{amsmath}
\usepackage{graphicx}
\usepackage{url}

\usepackage{tabularx}
\usepackage{multirow}
\usepackage{amssymb}

\usepackage{mdwmath}
\usepackage{mdwtab}
\usepackage{enumerate}

\usepackage{subfigure}
\usepackage{float}
\usepackage{tikz}
\usetikzlibrary{spy} 

\title{Learning Multi-Layer Transform Models}

\name{Saiprasad~Ravishankar and~Brendt~Wohlberg} 

\address{\hspace{0.0in} Theoretical Division, Los Alamos National Laboratory, Los Alamos, NM, USA \vspace{-0.04in}}

\begin{document}
%
\maketitle

\begin{abstract}
Learned data models based on sparsity are widely used in signal processing and imaging applications. A variety of methods for learning synthesis dictionaries, sparsifying transforms, etc., have been proposed in recent years, often imposing useful structures or properties on the models. In this work, we focus on sparsifying transform learning, which enjoys a number of advantages. We consider multi-layer or nested extensions of the transform model, and propose efficient learning algorithms. Numerical experiments with image data illustrate the behavior of the multi-layer transform learning algorithm and its usefulness for image denoising. Multi-layer models provide better denoising quality than single layer schemes.
\end{abstract}
\vspace{-0.15in}
\section{Introduction} \label{sec1}

\input{introduction_v1}

\vspace{-0.15in}
\section{Model and Algorithm} \label{sec2}

\input{modelandalgorithm_v1}

\vspace{-0.1in}
\section{Experiments} \label{sec3}

\input{experiments_v1}

\vspace{-0.1in}
\section{Conclusions} \label{sec4}
This paper investigated the learning of a multi-layer extension of the transform model, where the transform domain residuals generated in each layer were further sparsified in the subsequent layer. Filters in later layers were learned to jointly sparsify the coefficient residual maps of the preceding layers. We presented a greedy algorithm for the learning problem with a unitary constraint for the filters in each layer that enabled efficient filter updates. We also proposed downsampling the residual volumes in each layer to address dimensionality issues and to prevent overfitting to noisy data. Numerical experiments showed the promise of the learning algorithm in extracting rich image features and its utility for denoising by learning directly on noisy images. A simple decoder was used for the multi-layer model under the unitary assumption. The denoising quality typically improved with more layers and with multiple passes of denoising.
Future work will further explore the proposed model in detail with applications in inverse problems such as in imaging.

\bibliographystyle{IEEEbib}
\footnotesize{\bibliography{Multilayertransformlearning}}

\end{document}

%% file: introduction_v1.tex
Signal models based on sparsity, convolutional properties, or tensor or manifold structures, etc., have garnered increasing interest in recent years. Such models have been used in many applications including inverse problems, where they are often used to construct regularizers.
In particular, the learning of signal models from training data, or even corrupted measurements has shown promise in various settings.

Among sparsity-based models, the synthesis dictionary model \cite{ambruck} is perhaps the most well-known. Various methods have been proposed to learn synthesis dictionaries from signals or image patches \cite{elad, elad5, elad6, Yagh, zibul, Mai, sravfes} or in a convolutional framework \cite{wohlberg, garcbrendt}. However, the sparse coding problem (i.e., representing a signal as a sparse combination of appropriate dictionary atoms or filters) in the synthesis model (or during learning) typically lacks a closed-form solution and can be NP-hard in general. While numerous algorithms exist for general synthesis sparse coding \cite{pati, chen2,  befro, Needell2, wei}, they may be computationally expensive in large-scale settings.

On the other hand, several recent works \cite{sabres, sravbresov, sbclsTS2, doubsp2l, saiwen, Cai201489} have focused on the learning of sparsifying transform models, where the signal is assumed approximately sparse in a transform domain. A main advantage of the transform model is that the sparse approximation or sparse coding problem has a simple and efficient closed-form solution by thresholding. Adaptive sparsifying transforms have demonstrated promising performance in applications such as image and video denosing and medical image reconstruction \cite{sravTCI1, wenlibre, websaibr}. They can be learned relatively cheaply and the learning algorithms often come with provable convergence properties \cite{sbclsTS2, saiwen, saianadean}.
When the sparsifying transform is learned from or applied to regular patches of an image, the process involves convolutional/filtering operations. Recent works have thus interpreted transform learning for images as learning convolutional filters or filter banks \cite{pfisbres, yeravlongfes, sravancfes}.
Dictionary or transform learning methods often employ various structures, constraints, or regularizers for the model \cite{barchi1,sbclsTS2, doubsp2l, sravbresov, sravmoorrahfes} that may help avoid ambiguities, or make the model more efficient to use, or provide added robustness in applications, etc.

In this work, we focus on the sparsifying transform model and investigate a framework for learning multi-layer transforms. In each layer, the transform domain 
\emph{residuals} generated in the previous layer are further sparsified. Transforms are learned to jointly sparsify the residual maps (or effectively the residual volume). We optimize the proposed multi-layer transform learning problem in a greedy fashion by starting with the first or base layer and then estimating subsequent layers. A unitary constraint is used for the filters in each layer for simplicity, which leads to efficient alternating updates of the filters and coefficient maps. The residual maps are also undersampled in each layer (i.e., a subset is zeroed out) during learning, which helps avoid dimensionality issues and helps prevent overfitting to noise.
We also present a simple decoder to estimate the image from the multi-layer coefficients.

Our experiments illustrate the behavior of the proposed learning algorithm and the structure of the learned models and also demonstrate their use in image denoising, where multi-layer learned models outperform their single layer counterparts as well as conventional methods such as learned K-SVD dictionary-based denoising \cite{elad2}.
Finally, we also present a multi-pass extension of the learning scheme involving stacking multiple encoders and decoders that provides further improvements in denoising.
In the following sections, we discuss the proposed model and its learning along with the experiments on images.

%% file: modelandalgorithm_v1.tex
Here, we describe the transform model and its multi-layer or deep extension, and then present an optimization problem for its learning, together with a greedy algorithm for minimizing it.

\subsection{Multi-Layer Model and Learning Formulation}

The transform model suggests  that for a given signal $y \in \mathbb{R}^{n}$ and sparsifying transform operator $W \in \mathbb{R}^{m \times n}$, we have $Wy \approx z$, where $z$ has several zeros. Given the signal and operator, the best approximation $z$ is found cheaply by often thresholding $Wy$ with some threshold $\eta$, and setting the smallest elements (in magnitude) to zero \cite{sabres}.

\begin{figure*}[!t]
\begin{center}
\begin{tabular}{c}  
\hspace{-0.05in}\includegraphics[height=1.6in]{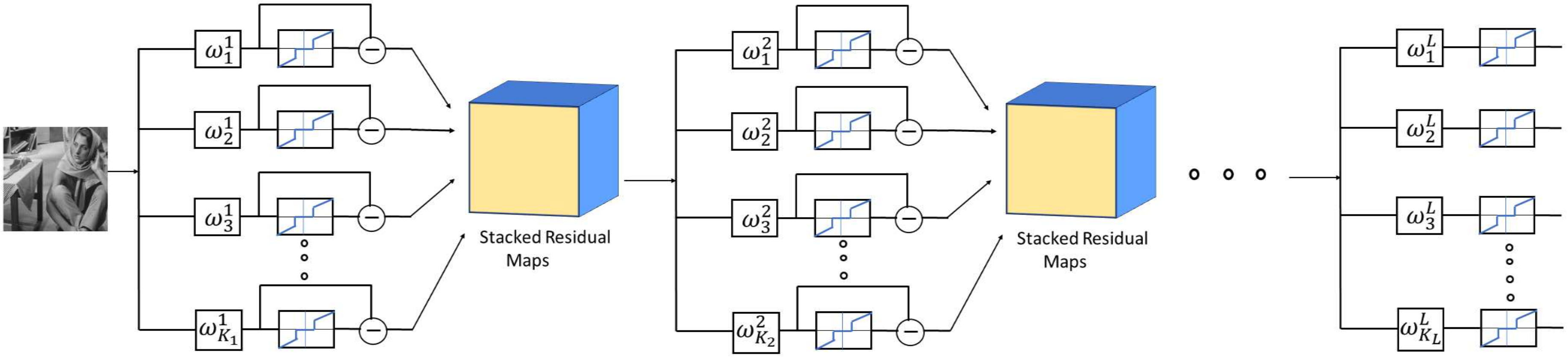}\\
\end{tabular}
\caption{The Deep Residual Transform (DeepResT) model with $L$ layers. The $l$th layer has $K_l$ filters and the filters in the second and higher layers sparsify the stacked residual maps (i.e., residual volumes). Note that residuals are not computed at the end of the $L$th (last) layer. The hard-thresholding function shown here can also be replaced with other non-linearities.}
\label{fig:model}
\end{center}
\end{figure*}

When the transform $W$ is applied to all overlapping patches (including patches overlapping image boundaries that wrap around on the other side of the image) of an image, each row or atom of the transform applies to all image patches via inner products to generate coefficient maps.
Clearly, this corresponds to a circular convolution (the corresponding filter is obtained by flipping and zero-padding a 
reshaped row of $W$) to generate each coefficient map, followed by thresholding.
When overlapping patches are used with a patch stride greater than $1$ pixel, applying the transform involves convolutions followed by downsampling and thresholding \cite{pfisbres}. 
Methods for learning such sparsifying transforms or filter banks from images typically enforce additional properties to avoid trivial solutions such as the all-zero operator or operators with repeated rows, etc.
Some of the regularizers employed in transform learning include $-\log | \det W| + \left \| W \right \|_F^2$ \cite{sabres}, or using a unitary constraint $W^{T}W=I$ \cite{sabres3}, or penalties enforcing incoherence \cite{sravbresov, pfisbres}.

We propose a multi-layer extension of the transform model that involves layers of sparsification. Fig.~\ref{fig:model} illustrates this model for $L$ layers. The filtering residuals (i.e., the difference between the pre- and post-thresholded coefficient map) or residual maps generated in the first layer for various filters are stacked together to form a residual volume. In the second and subsequent layers, the transform model jointly sparsifies the residual maps. For the $l$th layer, with $a_l  \times b_l \times c_l$ filters, we assume here that $c_l$ equals the residual volume depth\footnote{If the $c_l$ values were less than the residual volume depths and 3D filtering were performed in the second and higher layers, then the dimensions of the residual maps would keep increasing from one layer to another, potentially rendering the learning of filters infeasible.}, so that the convolution is done only along the spatial dimensions to produce 2D coefficient maps. In other words, the $c_l$ 2D filters/components are convolved with their corresponding 2D residual maps and the results are summed together to produce a 2D map.
In each layer in Fig.~\ref{fig:model}, the residual maps generated from different filters are stacked and further jointly sparsified in the next layer. However, in the final layer, only the sparse coefficient maps are computed without the residual maps.
We refer to this model as a Deep Residual Transform (DeepResT) model.
Intuitively, the residual maps in each layer may contain fine features/details, which could be further sparsified or encoded.

We now present an optimization framework to learn DeepResT models from images. The training images are denoted by the set $
\begin{Bmatrix}
x^{i}
\end{Bmatrix}$, with each $x^{i}$ a vectorized image.
To simplify the learning, we assume that the filters in each layer form a unitary set \cite{sabres3}, i.e., the matrix formed with the vectorized filters as its rows is unitary.
We formulate a patch-based learning problem (which may be equivalently written using convolutions) as follows:
\begin{align} 
\nonumber \mathrm{(P1)} \;\; & \min_{\begin{Bmatrix}
\Omega^{l}, Z^{l}
\end{Bmatrix}} \begin{Vmatrix} \Omega^{L}P^{L}(R^{L-1})-Z^{L} \end{Vmatrix}_{F}^{2} + \sum_{l=1}^{L}\eta_{l}^{2}\left \| Z^{l} \right \|_{0} \\
\nonumber & \;\;\;\; \text{s.t.} \;\, R^{j}=\Omega^{j}P^{j}(R^{j-1})-Z^{j}, \; 1\leq j \leq L-1,\\
\nonumber & \;\;\;\;\;\;\;\;\; \begin{pmatrix}
\Omega^{l}
\end{pmatrix}^{T} \Omega^{l} = I, \; 1 \leq l \leq L.
\end{align}
Here, $\begin{Bmatrix}
\Omega^{l}
\end{Bmatrix}_{l=1}^{L}$ denotes a set of $L$ unitary matrices, one for each layer, with $I$ denoting the identity matrix of appropriate dimensions.
The operator $P^{l}$ for each $1\leq l \leq L$ forms a matrix by extracting appropriately sized patches of its input and stacking them as vectorized matrix columns.
The matrices $\begin{Bmatrix}
Z^{l}
\end{Bmatrix}_{l=1}^{L}$ denote the sparse coefficient maps for the $L$ layers. Each row of $Z^{l}$ denotes the (row-vectorized) coefficient map\footnote{This could be multiple coefficient maps (for a filter) corresponding to multiple training images that are vectorized and stacked along the row.} for a particular atom or filter of $\Omega^{l}$. The $\ell_0$ ``norm" counts the total number of non-zero elements in a matrix or vector and the non-negative parameters $\begin{Bmatrix}
\eta_{l}
\end{Bmatrix}_{l=1}^{L}$ control the sparsity in each layer during training. While we use $\ell_{0}$ sparsity penalties, they could alternatively be replaced with $\ell_{1}$ or other sparsity penalties or constraints.
The residual maps are recursively defined by the matrices $R^{j}=\Omega^{j}P^{j}(R^{j-1})-Z^{j}$, where $R^{0}$ denotes the initial training images.

Problem (P1) is to learn the transforms for the $L$ layers by minimizing the norm of the residual in the $L$th layer (output) and enforcing the coefficient maps in the $L$ layers to be sparse via $\ell_{0}$ penalties. The learning in (P1) is quite different from the growing field of deep learning \cite{yanndeepl}, where the learning is typically supervised and objectives are task-driven (e.g., classification accuracy) rather than model-based (such as dictionary or transform learning costs). This enables utilizing (P1) to learn deep models from even corrupted data, without requiring large ground truth datasets for training. 
One could control the degrees of freedom while learning DeepResT models to achieve the best trade-offs in applications.

\textbf{Decoder.} The DeepResT model learned using (P1) acts as an encoder for images. In order to estimate an image(s) from the coefficient and residual maps, the residual and sparse coefficient estimates need to be backpropagated through the layers as follows. First, we obtain the following estimate from the $L$th layer model and coefficients:
\begin{equation} \label{RL}
P^{L}(\hat{R}^{L-1}) =\begin{pmatrix}
\hat{\Omega}^{L}
\end{pmatrix}^{T} \hat{Z}^{L}
\end{equation}
Then, the preceding residuals are computed one by one with decreasing $j$ as follows:
\begin{equation} \label{Rj}
P^{j}(\hat{R}^{j-1})= \begin{pmatrix}
\hat{\Omega}^{j}
\end{pmatrix}^{T}\hat{Z}^{j} +  \begin{pmatrix}
\hat{\Omega}^{j}
\end{pmatrix}^{T} \hat{R}^{j}, \;\; 1 \leq j \leq L-1.
\end{equation}
These updates follow quite easily under the unitary assumption for the transform matrices and the constraints in (P1). 
Each residual volume is obtained from its patch-version $P^{j}(\hat{R}^{j-1})$ by averaging together the patches at their respective locations in the volume (or image in the case of $j=1$) \cite{elad2, doubsp2l}. The estimated residual volume(s) are then reshaped in appropriate matrix form yielding $\hat{R}^{j-1}$.
In general, if the $\Omega^{j}$'s were non-unitary, the $\begin{pmatrix}
\hat{\Omega}^{j}
\end{pmatrix}^{T}$ operator in \eqref{RL} and \eqref{Rj} could be replaced with $\begin{pmatrix}
\hat{\Omega}^{j}
\end{pmatrix}^{\dagger}$, the pseudo-inverse.

Although there are $L$ sparsity parameters in (P1), we have observed in practice (see Section \ref{sec3}) that they can be set quite similarly and yet achieve good performance in applications.

\textbf{Downsampling Residual Volumes.} Another issue to consider for (P1) is the size of the transform in each layer. Assuming $\Omega^{1} \in \mathbb{R}^{n \times n}$, the residual volume depth after the first layer is $n$. Thus, the transform filters in the second layer (with their third dimension being $n$) will have size $\geq n$ upon vectorization. This implies that the transform matrix size would be monotonically increasing over the layers. 
In order to avoid the issue of increasing dimensionality and to achieve robustness to data noise and corruptions during learning, we propose a residual volume ``downsampling" strategy for each layer. The residual maps with the smallest energies are set to zero in each layer and not used to train the subsequent layer.
During the decoding process, these are simply stacked back as all-zero maps.

\subsection{DeepResT Learning Algorithm and Properties}

We propose a simple and fast greedy algorithm for (P1), where we learn the transform and coefficients one layer at a time.
At first, all the sparse codes $\begin{Bmatrix}
Z^{l}
\end{Bmatrix}_{l=1}^{L}$ are initialized to zero matrices.
Then, assuming that in each layer, patches are extracted such that each pixel in the residual volume (or original training images in the first layer) belongs to the same number of patches (e.g., with fully overlapping patches with patch wrap around), we have that
\begin{equation}\label{eq1}
\begin{Vmatrix}
\Omega^{j}P^{j}(R^{j-1})
\end{Vmatrix}_{F}^{2} = \begin{Vmatrix}
P^{j}(R^{j-1})
\end{Vmatrix}_{F}^{2} \propto \begin{Vmatrix}
R^{j-1}
\end{Vmatrix}_{F}^{2}
\end{equation}
Thus, when the initial sparse codes are all zero, $\begin{Vmatrix}
R^{j}
\end{Vmatrix}_{F}^{2} \propto \begin{Vmatrix}
R^{j-1}
\end{Vmatrix}_{F}^{2}$ $\forall$ $j$. Thus, minimizing sequentially with respect to $(\Omega^{l}, Z^{l})$ for $1 \leq l \leq L$ yields the following subproblems:
\begin{align} 
\nonumber \mathrm{(P2)} \;\; & \min_{\Omega^{l}, Z^{l}} \begin{Vmatrix} \Omega^{l}P^{l}(R^{l-1})-Z^{l} \end{Vmatrix}_{F}^{2} + \eta_{l}^{2}\left \| Z^{l} \right \|_{0} \\
\nonumber & \;\; \text{s.t.} \;\, \begin{pmatrix}
\Omega^{l}
\end{pmatrix}^{T}\Omega^{l} = I,
\end{align}
where for each $l$, $R^{l-1}$ is fixed based on the transforms and sparse coefficients estimated for the previous layers.

Problem (P2) is optimized by alternating between updating $\Omega^{l}$ and $Z^{l}$, with each subproblem solved efficiently \cite{sabres3}. With $\Omega^{l}$ fixed, the optimal coefficients are given as $\hat{Z}^{l} = H_{\eta_l}\begin{pmatrix}
\Omega^{l}P^{l}(R^{l-1})
\end{pmatrix}
$, where operator $H_{\eta_l}(\cdot)$ thresholds its inputs entry-wise by setting elements with magnitude less than $\eta_l$ to zero and leaving other elements unchanged. For fixed $Z^{l}$, the optimal operator $\hat{\Omega}^{l}$ is obtained as $VU^{T}$, where $U \Sigma V^{T}$ denotes the full singular value decomposition (SVD) of $P^{l}(R^{l-1})Z^{l^{T}}$.

When Problem (P1) is optimized by the above greedy algorithm with alternating optimization in each layer, the cost in (P1) decreases over the layers as well as within the alternating algorithm iterations in each layer.
When we downsample the residual volumes in each layer by keeping only a given number of residual maps or filter residuals, the operation could be included in (P1) by redefining $P^{l}$ to include downsampling followed by patch extraction. But \eqref{eq1} does not hold in this case. We still employ greedy sequential optimization based on (P2) to learn the model, which we found worked well in practice.

%% file: experiments_v1.tex
\begin{figure}[!t]
\begin{center}
\begin{tabular}{ccc}  
\includegraphics[height=0.9in]{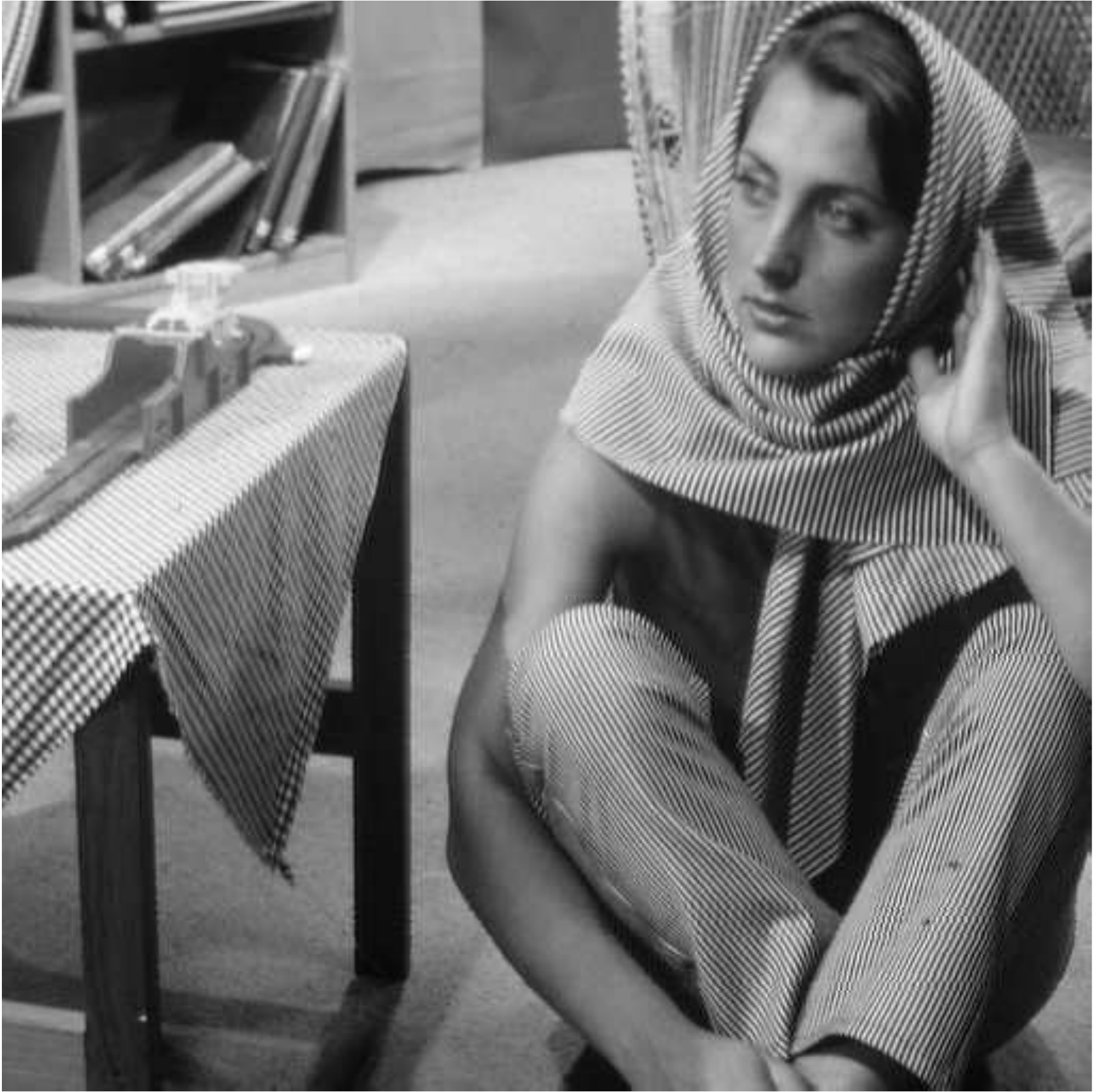} &
\includegraphics[height=0.9in]{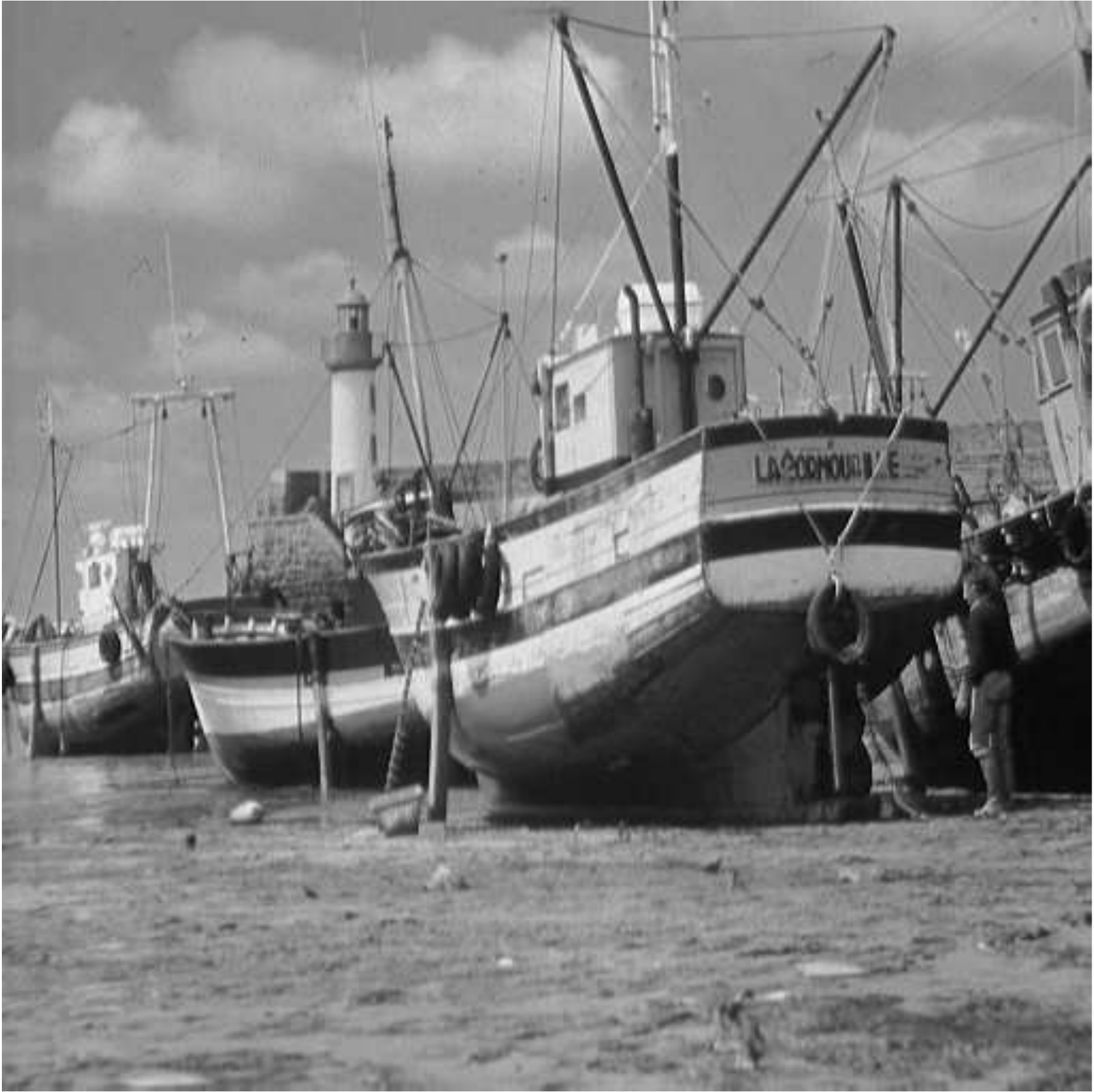} &
\includegraphics[height=0.9in]{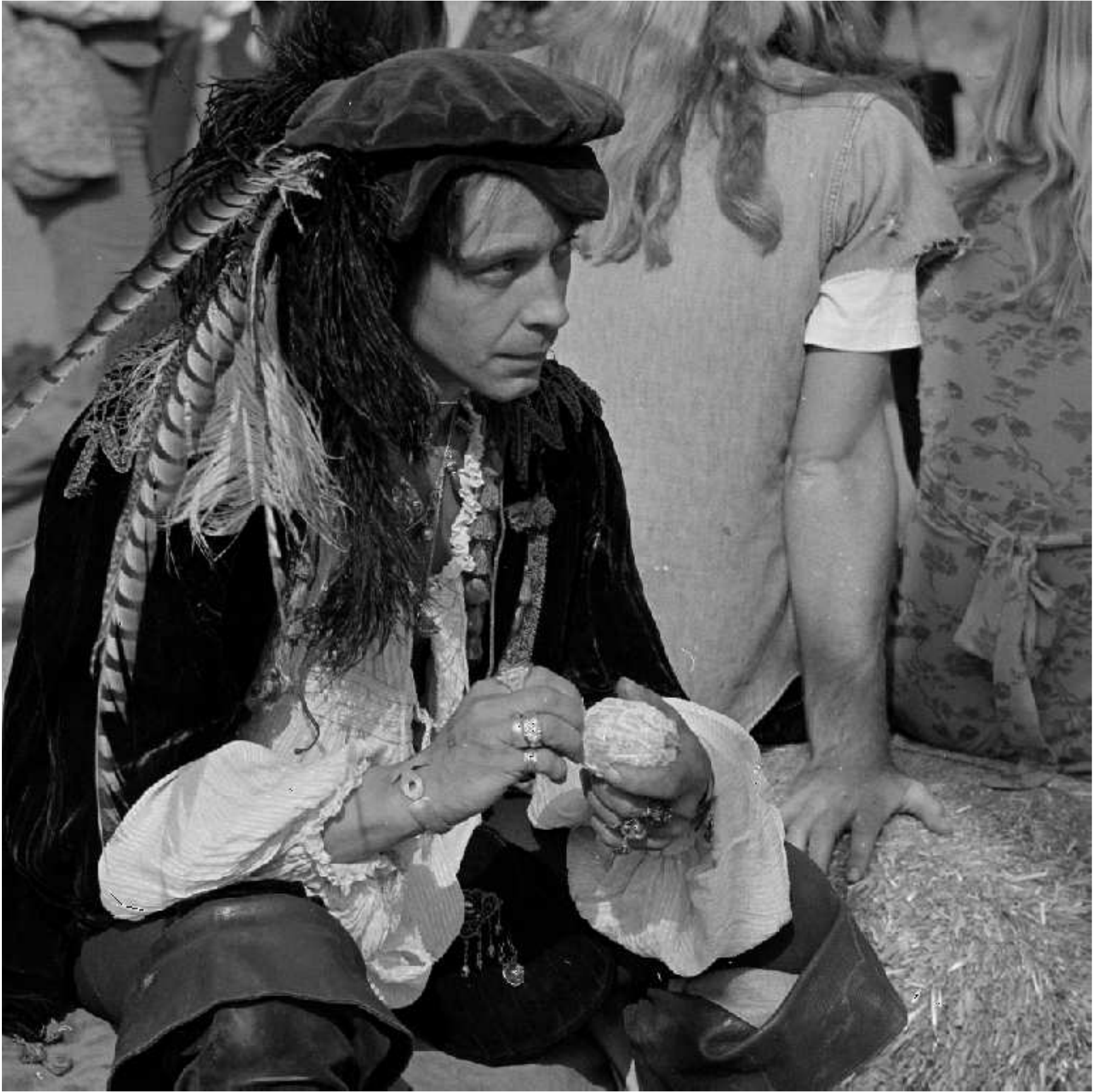}\\
\end{tabular}
\begin{tabular}{cc}
\includegraphics[height=0.9in]{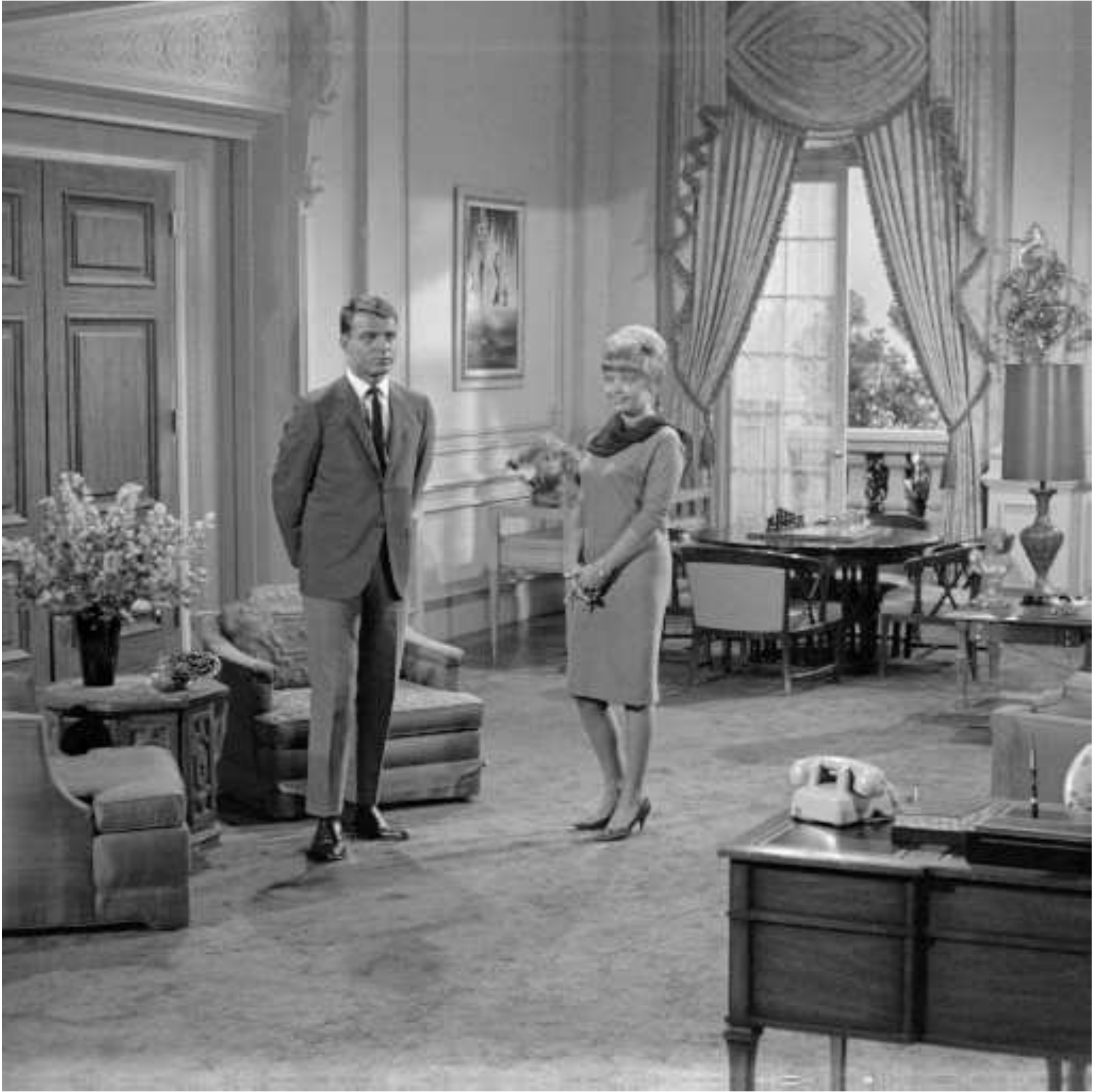}& 
\includegraphics[height=0.9in]{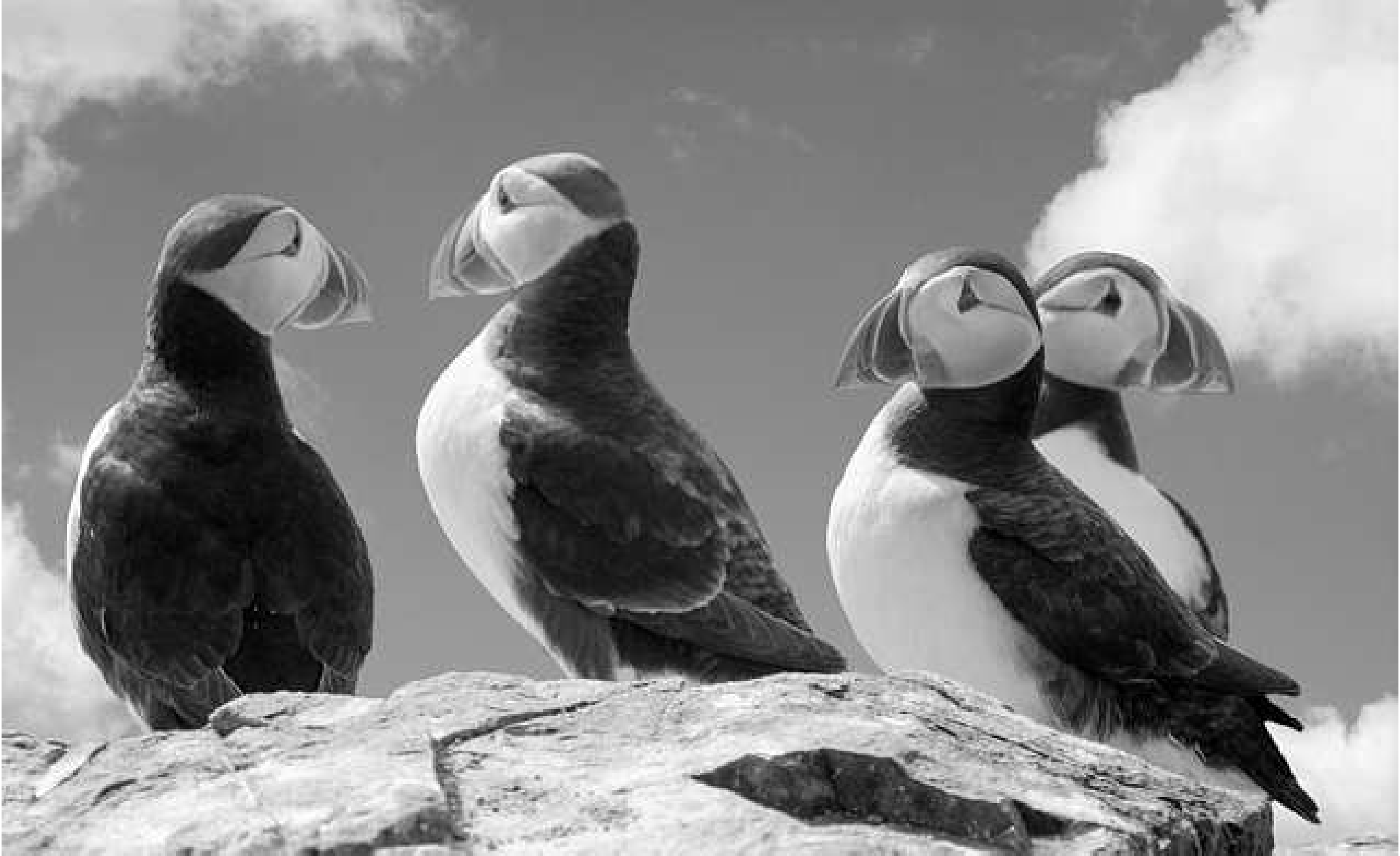}\\
\end{tabular}
\caption{Images used in experiments. Clockwise from top-left: Barbara ($512\times 512$), Boat ($512\times 512$), Man ($768 \times 768$), Puffins ($392 \times 640$), and Couple ($512\times 512$).}
\label{fig:testimages}
\end{center}
\vspace{-0.1in}
\end{figure}

Here, we present experiments illustrating the learned models for images and the behavior of the proposed learning algorithm for denoising. We refer to the proposed scheme as DeepResT. 
The images used in our experiments are shown in Fig.~\ref{fig:testimages}. 


\vspace{-0.1in}
\subsection{Multi-Layer Transforms for Images}

\begin{figure}[!t]
\begin{center}
\begin{tabular}{ccc}  
\includegraphics[height=1.15in]{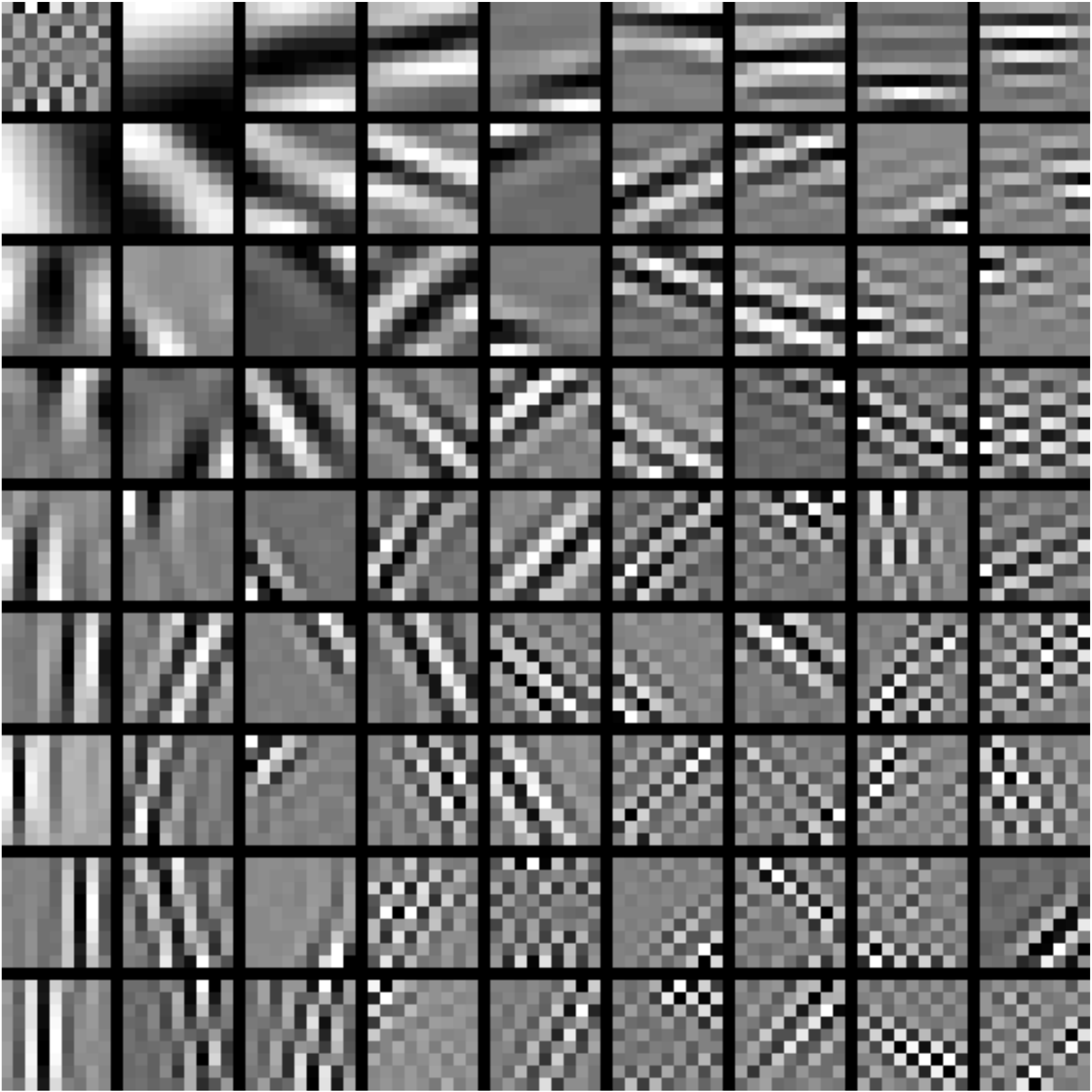} &
\hspace{-0.1in}\includegraphics[height=1.03in]{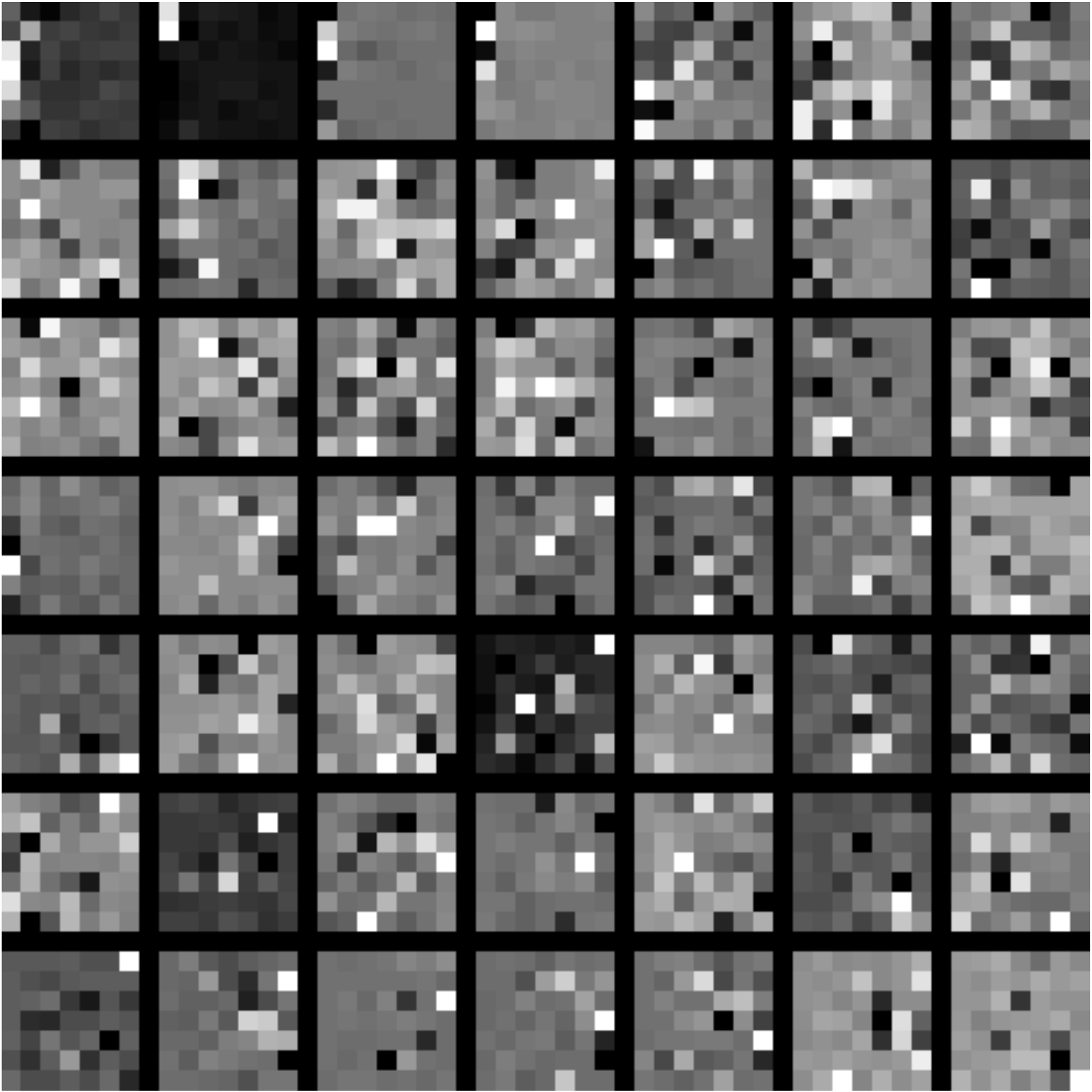} &
\hspace{-0.1in}\includegraphics[height=0.9in]{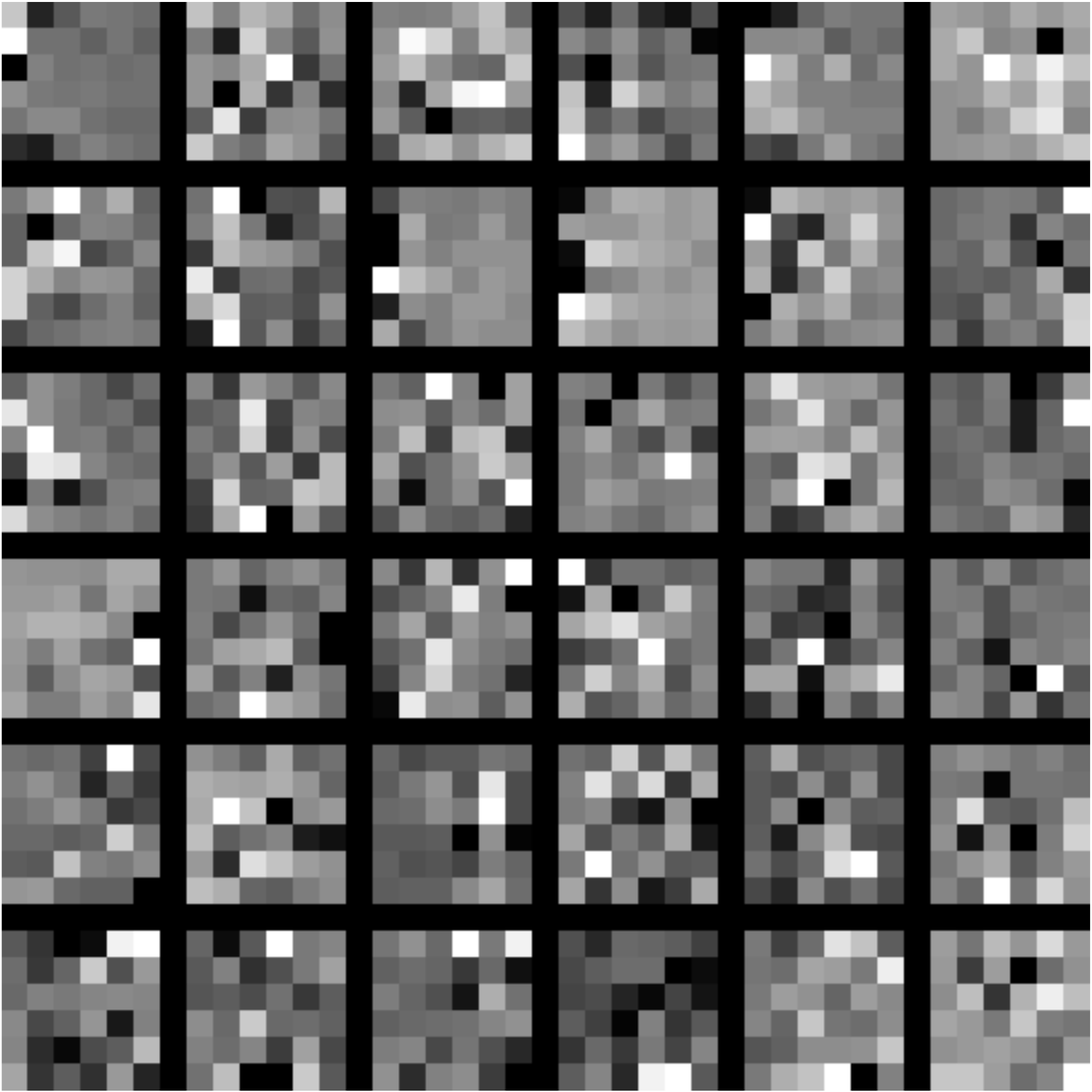}\\
\end{tabular}
\caption{Left to right: Transforms learned in the first, second, and third layers for the image Puffins. The atoms in each layer are shown as square 2D patches for concise display.}
\label{fig:models}
\end{center}
\vspace{-0.15in}
\end{figure}

We learned a DeepResT model with $L=3$ layers for the image Puffins. The patch size in the first layer was $9\times 9$, and in the second and third layers it was $1 \times 1 \times 49$ and $1 \times 1 \times 36$, respectively (i.e., $32$ and $13$ residual maps respectively were zeroed out for the second and third layers with the transform in these layers applied along the residual volume depth). We explored 1D filters in the higher layers in this work and leave the study of 3D filters to future work.
The threshold parameters were set as $\eta_{1} = 46.2$ and $\eta_2 = \eta_3 = 43.4$, and the greedy learning algorithm was executed with $400$ iterations in each layer. The initial transform was set to the 2D DCT in the first layer and the identity matrix in subsequent layers.

Fig.~\ref{fig:models} shows the transforms learned in each layer. The atoms are displayed as square patches. While the atoms in the first layer show directional and edge like features that sparsify the image, the 1D atoms in the second and third layers indicate how the different residual maps (arising from distinct filters) input to those layers were combined for better sparsification. The latter atoms clearly look quite different from the former image-level sparsifying features.
The benefit of the learned DeepResT over a single layer model is demonstrated next.

\vspace{-0.1in}
\subsection{Application to Image Denoising}

\begin{table}[t]
\centering
\begin{tabular}{|c|c|c|c|c|c|}
\hline
\multicolumn{1}{|c|}{Image} & \multicolumn{1}{c|}{$\sigma$} & \multicolumn{1}{c|}{K-SVD} & \multicolumn{3}{c|}{DeepResT} \\ 
\cline{4-6}
    &   &              &            $L=1$       &             $L=3$                &            $L=5$           \\
\hline
\multirow{4}{*}{Barbara} & 10 & 34.41  & 34.14 & \textbf{34.50} &  \textbf{34.50}\\ \cline{2-6}
 & 20 & 30.82 & 30.36 & \textbf{30.91}  &  \textbf{30.91} \\ \cline{2-6}
& 30 & 28.57   & 28.25  & \textbf{28.78}  & \textbf{28.78}\\ \cline{2-6}
 & 100 & 21.87  & 22.56 &  \textbf{22.71} &  22.67\\ 
\hline
\multirow{4}{*}{Boat} & 10 & 33.62 & 33.19 & 33.67 &  \textbf{33.72}\\ \cline{2-6}
 & 20 & 30.37 & 29.99 & 30.49  &  \textbf{30.54} \\ \cline{2-6}
& 30 & 28.43  & 28.16 & 28.66  & \textbf{28.70}\\ \cline{2-6}
 & 100 & 22.81 & 23.18 & \textbf{23.27}  &  23.26\\ 
\hline
\multirow{4}{*}{Man} & 10 & 32.73 & 32.34 & 32.84  &  \textbf{32.91}\\ \cline{2-6}
 & 20 & 29.40 & 29.07  & 29.66  &  \textbf{29.73} \\ \cline{2-6}
& 30 & 27.61  & 27.38 & 27.94  & \textbf{28.01}\\ \cline{2-6}
 & 100 & 22.75  & 23.17  & \textbf{23.28}  &  23.25\\ 
\hline
\multirow{4}{*}{Couple} & 10 & 33.51  & 33.15  & 33.59  &  \textbf{33.64}\\ \cline{2-6}
 & 20 & 30.03 & 29.72 & 30.22  &  \textbf{30.27} \\ \cline{2-6}
& 30 & 27.87  & 27.69  & 28.19  & \textbf{28.23}\\ \cline{2-6}
 & 100 & 22.57 & 22.88 & \textbf{22.99}  &  22.97\\ 
\hline
\multirow{4}{*}{Puffins} & 10 & 34.76 & 34.36 & 34.81 &  \textbf{34.85}\\ \cline{2-6}
 & 20 & 31.16 & 30.69 & 31.21 &  \textbf{31.24} \\ \cline{2-6}
& 30 & 29.18  & 28.71  & 29.23  & \textbf{29.26}\\ \cline{2-6}
 & 100 & 23.60  & 23.90  & \textbf{23.97}  &  23.92\\ 
\hline
\end{tabular}
\caption{PSNR values (in dB) for denoising for the adaptive transform algorithm using $L=1$, $L=3$, and $L=5$ layers. The denoising PSNRs obtained using the overcomplete K-SVD denoising scheme \cite{elad2} are also listed. The best PSNRs are marked in bold.}
\label{tab1}
\vspace{-0.05in}
\end{table}

We evaluate the usefulness of the adaptive DeepResT algorithm for denoising the images in Fig.~\ref{fig:testimages}. Simulated i.i.d. zero mean Gaussian noise with standard deviation $\sigma = 10, 20, 30$, and $100$ was added to the images. DeepResT learning is simulated with $L=1,3$, and $5$ layers.
The filter sizes in the layers were set as $9 \times 9$, $1 \times 1 \times 49$, $1 \times 1 \times 36$, $1 \times 1 \times 25$, and $1 \times 1 \times 16$, with appropriate numbers of low-energy residual maps zeroed while learning each layer. At the large $\sigma=100$, slightly smaller transform atom sizes were used for layers $2$ to $5$ to avoid noise overfitting, with the atom length along the third dimension being $36$, $25$, $16$, and $9$, respectively. The sparsity thresholds in the first and subsequent layers were set as $3.3 \sigma$ and $3.1 \sigma$, respectively, and the greedy training was run for $100$ iterations in each layer with the DeepResT model being learned from the noisy image and then used to denoise the same image.

\begin{table}[t]
\centering
\fontsize{8}{10pt}\selectfont
\begin{tabular}{|c|c|c|c|c|c|}
\hline
  & Barbara & Boat & Man & Couple & Puffins \\
\hline
Single Pass & 22.67   & 23.26   & 23.25    & 22.97    & 23.92   \\
\hline
Two Passes & 23.00   & 23.60  & 23.39   &  23.26   &  24.29  \\
\hline
\end{tabular}
\caption{Denoised PSNR values (in dB) for the adaptive DeepResT algorithm using a single pass and two passes with $L=5$ layers and $\sigma=100$.}
\label{tab2}
\end{table}

\begin{figure}[!t]
\begin{center}
\begin{tabular}{cc} 
\hspace{-0.1in}\includegraphics[height=2.28in]{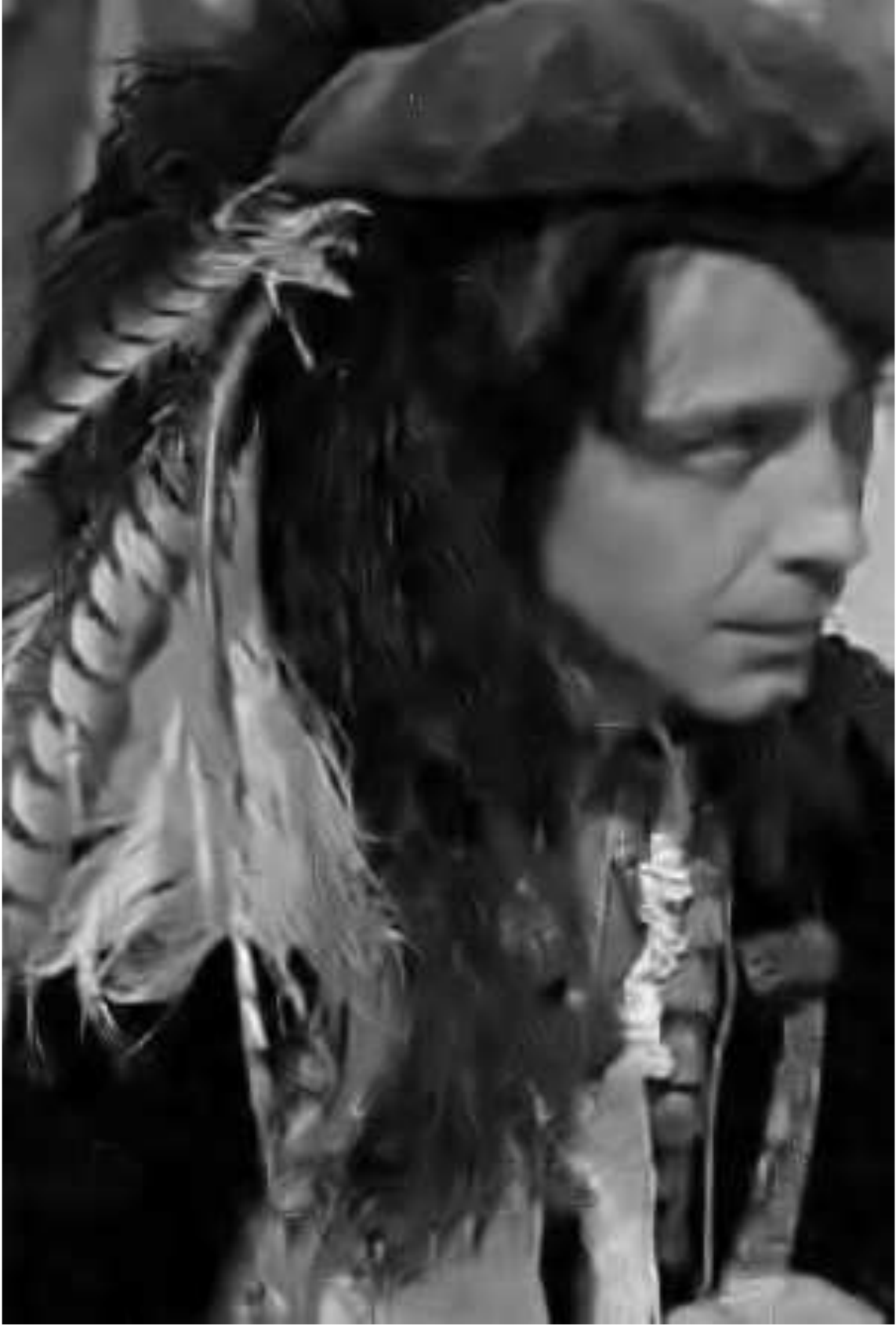} &
\hspace{-0.1in}\includegraphics[height=2.28in]{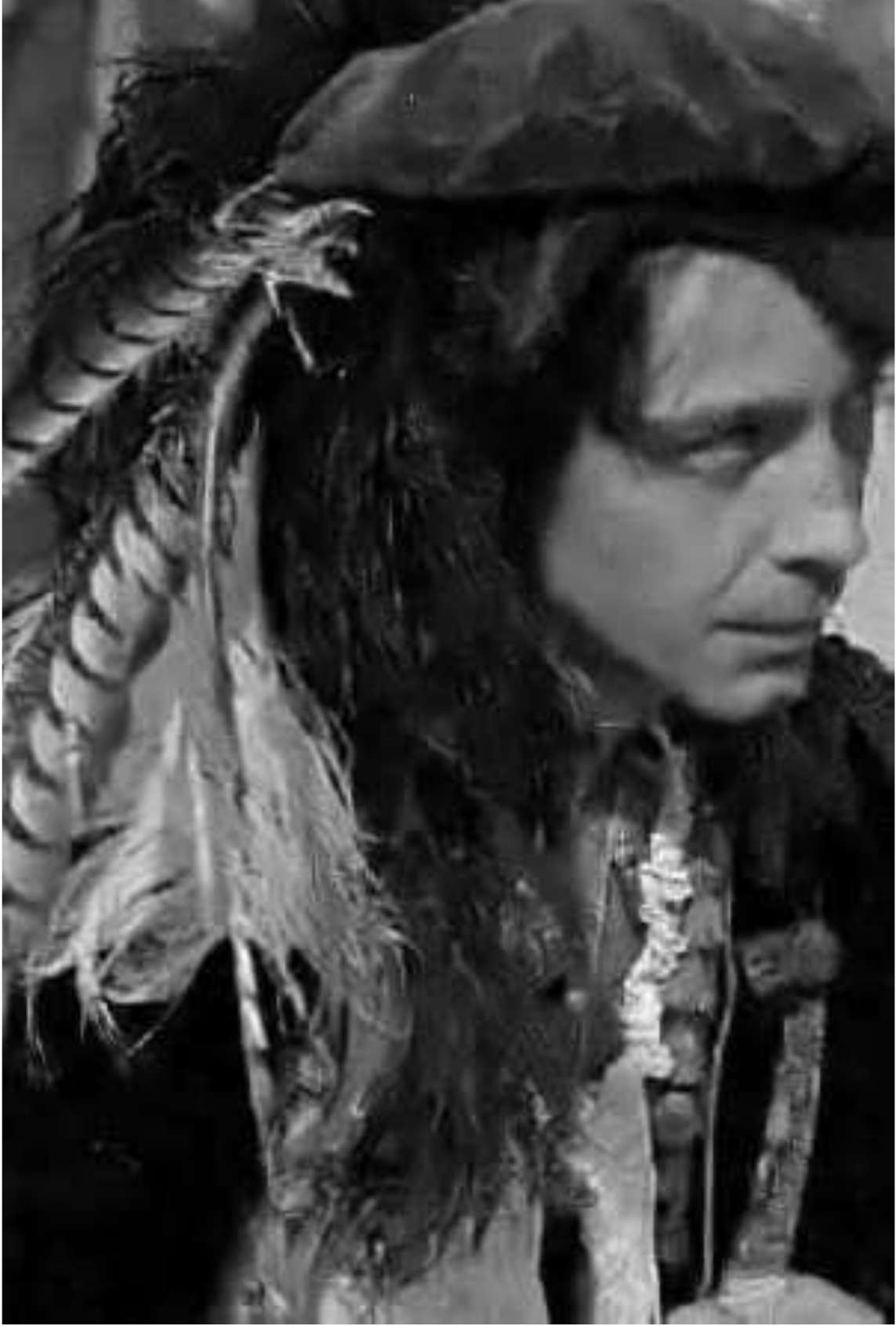}\\
\end{tabular}
\caption{
A zoomed-in region of the denoised image Man obtained using the single layer ($L=1$) adaptive transform scheme (left) and the DeepResT scheme with $L=5$ (right) when $\sigma=30$.}
\label{fig:denoising1}
\end{center}
\vspace{-0.15in}
\end{figure}

Table~\ref{tab1} shows the peak signal-to-noise-ratio (PSNR) values in decibels (dB) for adaptive transform denoising with various numbers of layers along with the PSNRs for denoising with the well-known overcomplete K-SVD learned dictionary-based denoising algorithm \cite{elad2}. 
The DeepResT method with $L=5$ and $L=3$ layers perform quite similarly (with the former outperforming by 0.02 dB on average), and both outperform the K-SVD method and the single-layer transform learning-based denoising scheme. In particular, the $L=5$ case outperforms K-SVD by 0.26 dB on average, with a peak improvement of 0.8 dB.
Note that the K-SVD method sparse codes patches according to an $\ell_2$ error bound ($\propto \sigma$) criterion, which plays a key role in its success. Incorporating such a bound into our scheme (see \cite{doubsp2l} for its use in single-layer adaptive transform denoising) may improve its performance further over the current sparsity penalized strategy.
Fig.~\ref{fig:denoising1} shows the zoom-ins of the denoised image Man with $L=1$ and $L=5$ layers. The image with $5$ layers shows sharper edges than the one with the single layer scheme.
Finally, when DeepResT models with $5$ layers and with $8 \times 8$ or length-$64$ atoms in the first layer and smaller atoms for subsequent layers were learned to adaptively denoise images, the PSNR values were only $0.07$ dB worse on average than for the $5$ layer models in Table~\ref{tab1}.



\begin{figure}[!t]
\begin{center}
\begin{tabular}{cc}  
\hspace{-0.1in}\includegraphics[height=2.38in]{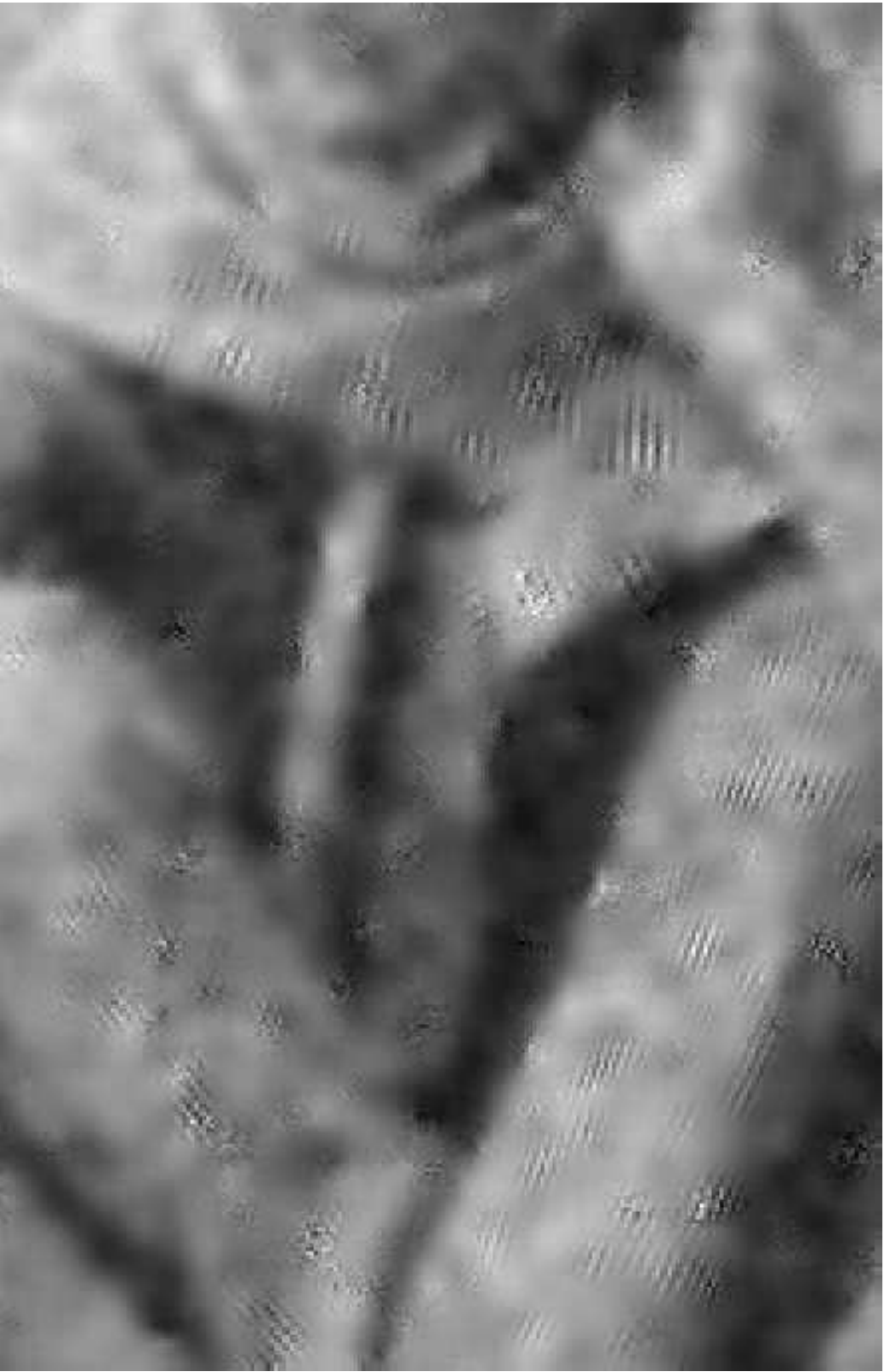} &
\hspace{-0.1in}\includegraphics[height=2.38in]{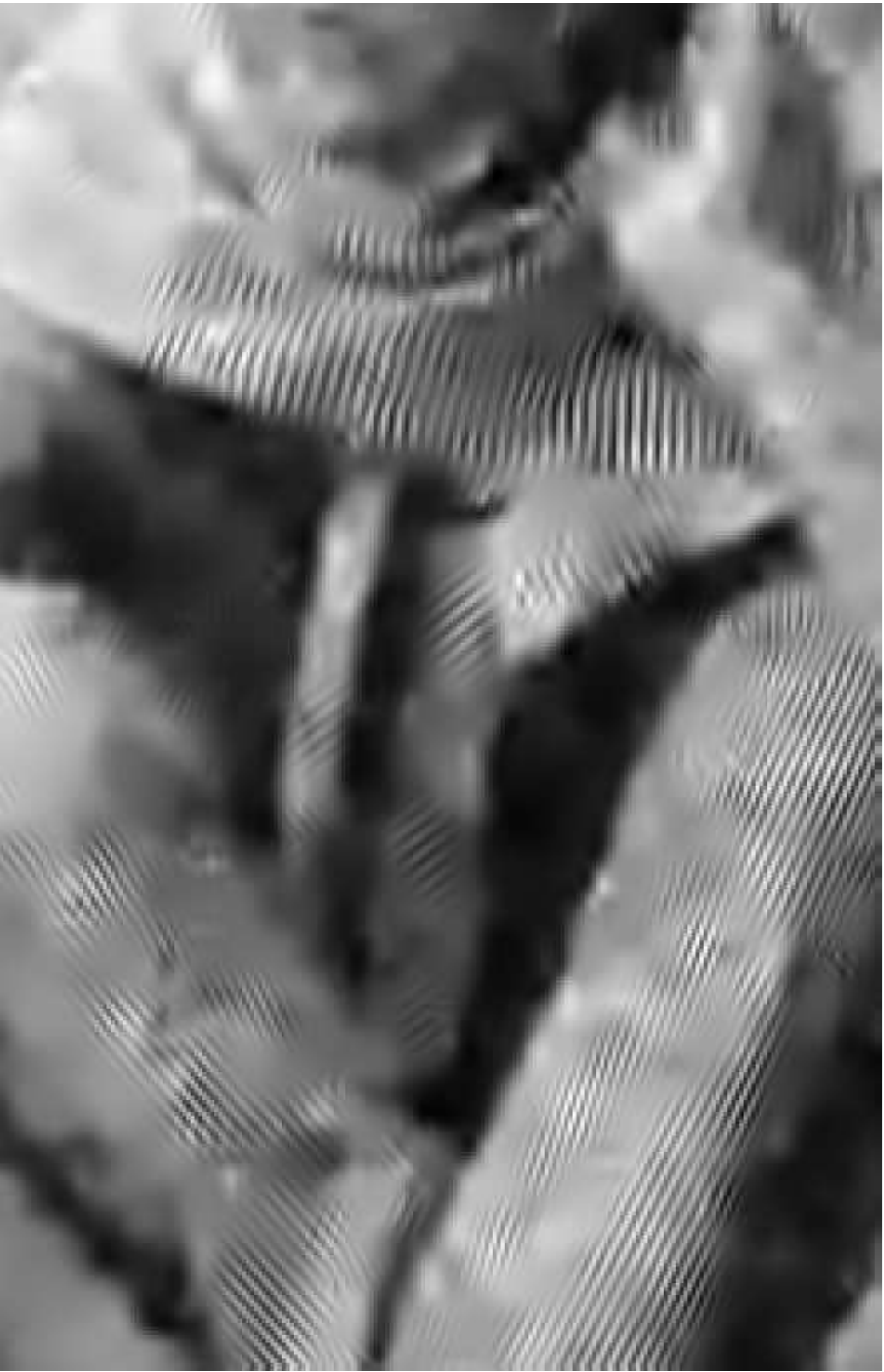}\\
\end{tabular}
\caption{
Zoomed-in region of the denoised image Barbara, obtained using K-SVD (left) and the DeepResT method with $L=5$ and two passes (right) for $\sigma=100$.}
\label{fig:denoising2}
\end{center}
\vspace{-0.15in}
\end{figure}

\textbf{Multi-pass or Stacked DeepResT Scheme.}
We also studied a multi-pass version of the DeepResT scheme, where the image denoised by the learned DeepResT model is further denoised by additional passes of DeepResT learning. This corresponds to stacking several DeepResT encoder + decoder modules to perform denoising.
Table~\ref{tab2} shows the PSNR values with one and two passes of learned DeepResT ($L=5$) denoising at $\sigma=100$. For the two pass scheme, the $\sigma$ value that determines the thresholds in each pass was set as $90$ (a smaller estimate in the first pass could enable further denoising improvement in the next pass) and $20$ in the first and second pass, respectively.
The two pass scheme achieves a peak improvement of about 0.4 dB over the single pass scheme.
Fig.~\ref{fig:denoising2} shows zoom-ins of the denoised image Barbara with K-SVD denoising \cite{elad2} and the two-pass DeepResT scheme showing much better reconstruction of image features and textures for the latter approach.


Recent works \cite{wenlibre, websaibr} have shown that combining transform learning with block matching strategies can outperform popular state-of-the-art image and video denoising methods such as BM3D and VBM3D or VBM4D. The proposed DeepResT learning could also be potentially combined with block matching strategies. We leave its investigation to future work.